\documentclass[runningheads]{llncs}
\usepackage{graphicx}
\usepackage{subcaption}
\usepackage{booktabs}
\usepackage[hidelinks]{hyperref}
\usepackage{cite}
\usepackage{amsmath,amssymb,amsfonts}
\usepackage{algorithmic}
\usepackage{graphicx}
\usepackage{pdfpages}
\usepackage{multirow}
\usepackage{array}
\usepackage{caption}
\usepackage{float}
\usepackage{multirow}
\usepackage{enumitem}
\setlist{nosep}

\usepackage[T5]{fontenc}
\usepackage{balance}
\usepackage[utf8]{inputenc}
\DeclareTextSymbolDefault{\DH}{T1}
\usepackage{lmodern} 

\usepackage{textcomp}
\usepackage{xcolor}
\def\BibTeX{{\rm B\kern-.05em{\sc i\kern-.025em b}\kern-.08em
    T\kern-.1667em\lower.7ex\hbox{E}\kern-.125emX}}

\usepackage{orcidlink}

\newcolumntype{Y}{>{\centering\arraybackslash}X}



\makeatletter
\newcommand{\printfnsymbol}[1]{%
  \textsuperscript{\@fnsymbol{#1}}%
}
\makeatother

\begin{document}

\title{Forged Calamity: Benchmark for Cross-Domain Synthetic Disaster Detection in the Age of Diffusion}

\titlerunning{Forged Calamity: Benchmark for Cross-Domain Synthetic Disaster Detection}

\author{
Duc-Manh Phan\inst{1,2}\orcidlink{0009-0006-6648-3844}\thanks{These authors contributed equally to this work.} \and Quoc-Duy Tran\inst{1,2}\orcidlink{0009-0003-1773-1520}\printfnsymbol{1} \and Duy-Khang Do\inst{1,2}\orcidlink{0009-0003-7430-1586}\printfnsymbol{1} \and 
Anh-Tuan Vo\inst{1,2}\orcidlink{0009-0002-3547-9907} \and Hai-Dang Nguyen\inst{1,2}\orcidlink{0000-0003-0888-8908} \and Trong Le Do\inst{1,2}\orcidlink{0000-0002-2906-0360} \and Mai-Khiem Tran\inst{1,2}\orcidlink{0000-0001-5460-0229} \and Vinh-Tiep Nguyen\inst{2,3}\orcidlink{0000-0003-4260-7874} \and Tam V. Nguyen\inst{4}\orcidlink{0000-0003-0236-7992} \and Isao Echizen\inst{5}\orcidlink{0000-0003-4908-1860} \and
Minh-Triet Tran\inst{1,2}\orcidlink{0000-0003-3046-3041} \and
Trung-Nghia Le\thanks{Corresponding author.}\inst{1,2}\orcidlink{0000-0002-7363-2610}}

\institute{University of Science, VNU-HCM, Ho Chi Minh, Vietnam \and
Vietnam National University, Ho Chi Minh, Vietnam \and University of Information technology, VNU-HCM, Ho Chi Minh, Vietnam \and University of Dayton, Ohio, United States \and National Institute of Informatics, Tokyo, Japan\\
\email{\{22120205,22120082,22120454,22120406\}@student.hcmus.edu.vn},\\\email{\{nhdang, dtle, tmkhiem\}@selab.hcmus.edu.vn},\\\email{tiepnv@uit.edu.vn},\email{tamnguyen@udayton.edu},\email{
iechizen@nii.ac.jp},\\
\email{\{tmtriet, ltnghia\}@fit.hcmus.edu.vn}}

\authorrunning{Duc-Manh Phan et al.}

\maketitle

\begin{abstract}

The rapid advancement of text-to-image diffusion models has enabled the creation of highly photorealistic synthetic images that closely resemble real photographs, making it increasingly difficult to distinguish authentic content from AI-generated fabrications. This poses challenges for cybersecurity, digital forensics, and disaster response, where fake imagery of floods, fires, or earthquakes can spread misinformation or disrupt emergency operations. To address this, we introduce Forged Calamity, a benchmark dataset for synthetic disaster detection containing 30,000 images, including 6,000 real and 24,000 synthetic samples generated by four diffusion models. Comprehensive experiments across fine-tuned and zero-shot settings reveal consistent weaknesses in current forensic approaches. Fine-tuned detectors perform well in-distribution but lose up to 50\% accuracy on unseen generators or disaster types, showing overfitting to model-specific artifacts. Zero-shot generalized detectors also struggle to maintain stable accuracy, with only limited resilience in a few representation-robust models. These findings highlight persistent generalization gaps and the urgent need for domain- and model-agnostic detection methods to ensure visual authenticity in the diffusion era.

\keywords{
AI-generated imagery, Diffusion models, Synthetic disaster detection, Multimedia forensics, Cross-domain generalization.
}

\end{abstract}

\section{Introduction}

The rapid advancement of latent diffusion models (LDMs)~\cite{Rombach_2022_CVPR, podell2023sdxlimprovinglatentdiffusion, chen2024pixartsigma} has enabled the creation of photorealistic synthetic images that closely mimic real-world photographs. Trained on large-scale datasets and conditioned through textual prompts, these generative models have made it increasingly difficult to distinguish between authentic and AI-generated content. Unlike earlier Generative Adversarial Network (GAN)-based approaches~\cite{goodfellow2014generativeadversarialnetworks}, LDMs operate in a compressed latent space that allows efficient image synthesis while minimizing the visual artifacts commonly found in older methods.

This technological evolution presents new risks in cybersecurity and media forensics. Synthetic imagery depicting natural disasters, such as floods, fires, or earthquakes, can be weaponized to spread misinformation, disrupt emergency response systems, or manipulate public discourse. Verifying the authenticity of disaster-related imagery has therefore become essential for digital forensics, media verification, and national security infrastructures~\cite{Monahan_Martin_Zaitsev_Bartelt_RahmanNoordeen_2025}. Although human observers may notice subtle visual cues, such as unnatural textures or hyper-realistic color tones, these indicators vary widely across generative models, making manual detection unreliable.

In this work, we investigate the generalization capacity of deep learning classifiers when fine-tuned to distinguish between real and AI-generated disaster images. To support this study, we introduce Forged Calamity, a large-scale benchmark dataset for synthetic disaster detection. The dataset includes 6,000 real images from the Incident-1M dataset~\cite{weber2022incidents1mlargescaledatasetimages}, filtered into four categories: fire, earthquake, flood, and thunderstorm. For each category, we generate 6,000 synthetic images using four state-of-the-art diffusion models: Stable Diffusion 1.5~\cite{Rombach_2022_CVPR}, Stable Diffusion 2.0~\cite{Rombach_2022_CVPR}, Stable Diffusion XL~\cite{podell2023sdxlimprovinglatentdiffusion}, and PixArt~\cite{chen2024pixartsigma}. The resulting dataset comprises 30,000 images that provide a comprehensive benchmark across both semantic and generative domains.

Our experimental design simulates realistic forensic scenarios in which a detector is trained on one disaster type generated by a specific diffusion model (for example, fire from SD 1.5) and then evaluated on unseen disaster types and alternative generators. This setup allows us to assess both domain transferability (across disaster categories) and model transferability (across generative models), which are essential for building robust detection systems.

Our benchmarking results reveal fundamental weaknesses in the robustness of current vision-based detection systems. Models trained on a single diffusion generator perform well in-distribution but show significant accuracy degradation, often exceeding 40\%, when evaluated on unseen generators or new disaster categories. Fine-tuned detectors, despite leveraging large-scale pre-trained backbones, consistently overfit to the visual "fingerprints" of the source generator and fail to generalize to new architectures or data domains. Zero-shot evaluations reinforce this limitation, as most existing methods cannot maintain consistent performance without dataset-specific adaptation. These observations confirm that current detection systems are far from universal, struggling to adapt to the rapidly evolving diffusion landscape and underscoring the need for robust, generalizable solutions that can handle both semantic diversity and generative variability.

Our contributions are as follows:

\begin{itemize}
\item We present Forged Calamity, a large-scale dataset comprising 30,000 real and synthetic disaster images across four categories and four diffusion generators, establishing a dedicated benchmark for studying generalization in forensic detection.

\item We conduct extensive experiments under both fine-tuned and zero-shot conditions to evaluate model robustness across semantic and generative shifts

\item Results show that detectors achieve high in-distribution accuracy but fail to generalize across unseen diffusion models or disaster types, with accuracy drops of up to 50\% in challenging scenarios. The benchmark exposes critical limitations in current detection strategies, providing a foundation for developing more robust, domain-agnostic, and generalizable forensic approaches.

\end{itemize}

\section{Related Work}

\subsection{Disaster Image Datasets}

Research on disaster imagery has produced several benchmark datasets for scene understanding, damage assessment, and crisis management. For instance, CrisisMMD~\cite{alam2018crisismmd} collects multimodal Twitter data from various natural disasters and provides annotations for informativeness, humanitarian categories, and damage severity. The xBD dataset~\cite{gupta2019xBDbuildingdamage} offers pre- and post-event satellite imagery with building polygons and ordinal damage labels, enabling large-scale change detection and damage evaluation. Similarly, FloodNet~\cite{rahnemoon2020floodnet} contains high-resolution UAV imagery captured after Hurricane Harvey, with pixel-wise semantic labels and question–answer pairs designed for segmentation and visual question answering tasks.

While these datasets have significantly advanced the field of disaster informatics, their focus remains on classification, segmentation, and damage assessment. None address the emerging challenge of differentiating between authentic and AI-generated disaster imagery. This gap is critical because synthetic disaster images can propagate misinformation, hinder emergency response efforts, and erode public trust in visual evidence during crises.

\subsection{AI-Generated Image Detection}

The field of computer vision has developed several benchmarks for detecting AI-generated imagery. Our study builds upon this growing body of research, focusing on synthetic image detection and the persistent challenge of model generalization.

Early work in fake image detection primarily targeted artifacts produced by GANs~\cite{goodfellow2014generativeadversarialnetworks}, the first models capable of producing high-fidelity synthetic images. Detectors from this era relied on statistical inconsistencies in the frequency domain or the absence of camera-specific noise patterns, such as PRNU~\cite{frank2020leveragingfrequencyanalysisdeep, Verdoliva_2020}. Although effective against architectures like StyleGAN~\cite{karras2019stylebasedgeneratorarchitecturegenerative}, these methods were often specialized to the generation artifacts of specific GAN families.

The emergence of LDMs~\cite{Rombach_2022_CVPR}, has introduced a more formidable challenge. By operating in a compressed latent space and leveraging iterative denoising, LDMs generate photorealistic images with minimal structural artifacts, rendering many traditional detection strategies ineffective~\cite{dhariwal2021diffusionmodelsbeatgans}. As a result, models trained to detect GAN-generated images often fail when confronted with diffusion-generated content, and vice versa~\cite{corvi2022detectionsyntheticimagesgenerated}.

% To address scalability, GenImage~\cite{zhu2023genimage} was introduced as a million-scale benchmark for fake image detection. It defines two evaluation tasks, including cross-generator classification for assessing generalization to unseen generators, and degraded image classification for testing robustness under real-world distortions such as compression or blurring. While GenImage provides a strong foundation for large-scale fake image detection, it remains domain-agnostic and does not cover high-stakes contexts like disaster imagery, where misinformation can have severe social consequences.

This gap underscores a central challenge in modern detection research: generalization. Numerous studies have shown that even state-of-the-art detectors struggle when applied to unseen generative models or novel content domains~\cite{ojha2024universalfakeimagedetectors}. This limited robustness remains a major barrier to deploying practical detection systems. While recent work has explored data augmentation~\cite{zhao2021learningselfconsistencydeepfakedetection} and domain-agnostic representation learning~\cite{Guarnera2022} to improve resilience, a comprehensive understanding of how modern vision architectures generalize across both generative and semantic variations is still lacking.

Our work directly addresses this problem by systematically evaluating multiple vision backbones under a rigorous cross-model and cross-domain setting, using disaster imagery as a high-stakes and socially relevant testbed.

\section{Proposed Forged Calamity Dataset}

\subsection{Real-World Image Collection}

\begin{figure}[!t]
    \centering
    \includegraphics[trim={0 0 125mm 0},clip,width=\textwidth]{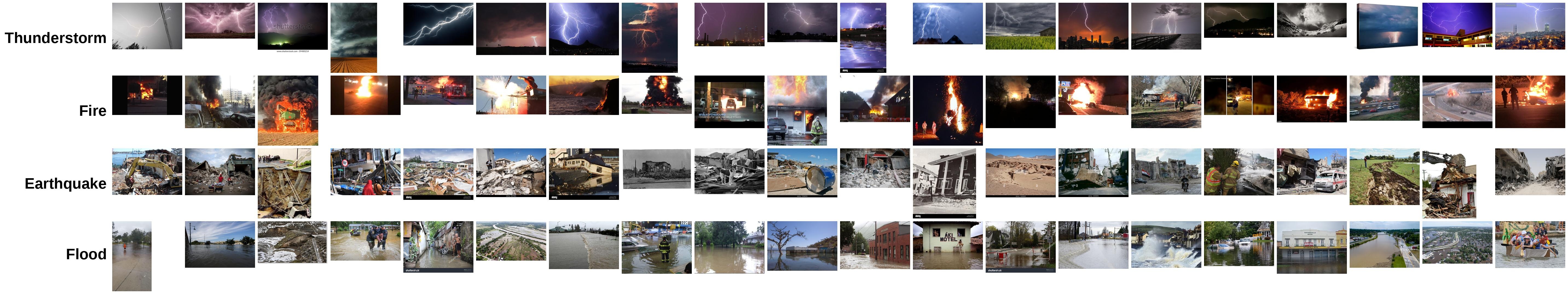}
    \caption{Representative examples of natural disaster categories from the Incident-1M dataset~\cite{weber2022incidents1mlargescaledatasetimages}.}
    \label{fig:real_images}
    \vspace{-5mm}
\end{figure}

The real-world image data used in this study was sourced from the Incident-1M dataset~\cite{weber2022incidents1mlargescaledatasetimages}, a large and diverse collection of disaster-related imagery, and from the Internet. A targeted subset was curated to represent four natural disaster categories: \textit{Fire}, \textit{Earthquake}, \textit{Flood}, and \textit{Thunderstorm}. For each category, 1,500 high-quality images were carefully selected, resulting in a total of 6,000 real images. This subset constitutes the ground-truth or “REAL” class. Representative samples from this curated collection are illustrated in Fig.~\ref{fig:real_images}.

\begin{figure}[!t]
    \centering
    \includegraphics[width=\textwidth]{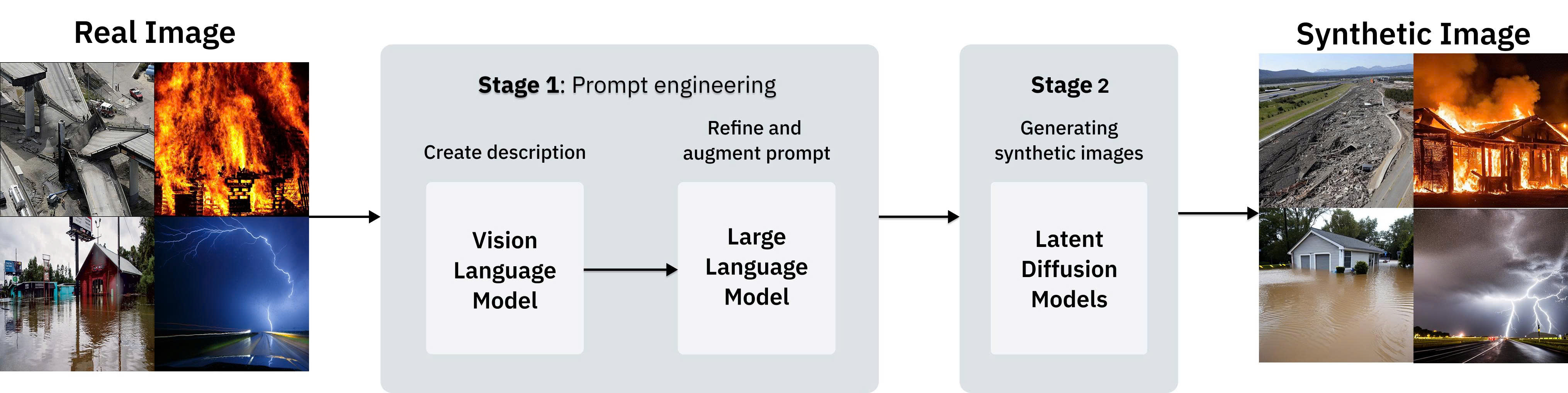}
    \caption{Overview of the synthetic image generation pipeline. The process consists of two main stages: (1) prompt engineering, in which detailed textual descriptions of disaster scenarios are created and refined for optimal input, and (2) image synthesis, in which multiple text-to-image diffusion models generate high-quality visual depictions based on the refined prompts.}
    \label{fig:pipeline-generated}
    \vspace{-5mm}
\end{figure}

\subsection{Synthetic Image Construction}

To investigate the characteristics and detectability of AI-generated imagery, we constructed a synthetic image dataset following the two-stage pipeline illustrated in Fig.~\ref{fig:pipeline-generated}. The pipeline comprises two main stages: {prompt engineering} and {image synthesis}.

\textbf{Prompt Engineering.}
The first stage focuses on generating descriptive textual prompts for each disaster category. This process involves two steps.  
\textit{Detailed prompt formulation:} The Moondream tool~\cite{vik_2024} was used to produce rich textual descriptions that capture fine-grained visual details associated with each disaster scenario.  
\textit{Prompt refinement:} These detailed descriptions were then processed using Llama~3~\cite{llama3modelcard} to distill them into concise, semantically coherent prompts optimized for text-to-image generation.

\textbf{Image Synthesis.}
With the refined prompts prepared, the second stage performs large-scale image synthesis. Each prompt was used as input for four state-of-the-art text-to-image diffusion models: Stable Diffusion~1.5~\cite{Rombach_2022_CVPR} (SD~1.5), Stable Diffusion~2.0~\cite{Rombach_2022_CVPR} (SD~2.0), Stable Diffusion~XL~\cite{podell2023sdxlimprovinglatentdiffusion} (SDXL), and PixArt~\cite{chen2024pixartsigma}. Each model generated 1,500 synthetic images for each of the four natural disaster categories, resulting in a diverse and comprehensive set of AI-generated samples for analysis. To ensure dataset diversity and eliminate redundancy, we adopted the similarity filtering method proposed by Nguyen et al.~\cite{nguyen2025automated} to remove highly similar generated images.

\subsection{Dataset Description}

\begin{figure}[!t]
    \centering
    \includegraphics[width=\linewidth]{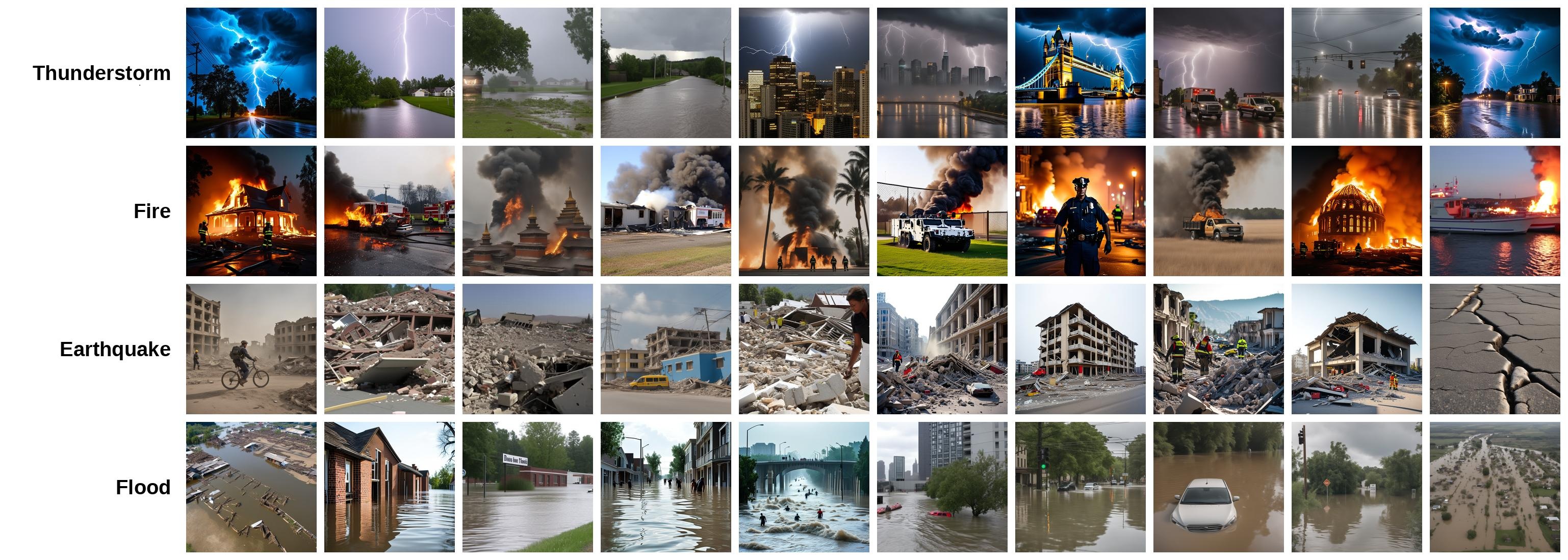}
    \caption{Examples of AI-generated images within our Forged Calamity dataset. Each image corresponds to one of the four disaster categories.}
    \label{fig:examples-synth}
    \vspace{-5mm}
\end{figure}

The Forged Calamity dataset comprises a total of 30,000 images, including 6,000 real and 24,000 synthetic samples (Table~\ref{tab:dataset}). Real images have 1,500 images for each of the four disaster categories: Fire, earthquake, Flood, and thunderstorm (Fig.~\ref{fig:examples-synth}). Synthetic counterparts were generated using four text-to-image diffusion models (e.g., Stable Diffusion 1.5, Stable Diffusion 2.0, Stable Diffusion XL, and PixArt), each producing 1,500 images per category. The generated images were organized into four subsets based on their source models (\texttt{SD 1.5}, \texttt{SD 2.0}, \texttt{SD XL}, and \texttt{PixArt}). In our  experiments, the synthetic images collectively form the negative class, providing a comprehensive and balanced benchmark for evaluating the ability of vision models to distinguish authentic disaster imagery from AI-generated fabrications.

\begin{table}[t!]
\centering
\caption{Composition of the Forged Calamity dataset with real and synthetic images generated across four disaster categories and four diffusion models.}
\label{tab:dataset}
\begin{tabular}{l|c|cccc}
\toprule
\multirow{2}{*}{\textbf{\begin{tabular}[c]
{@{}l@{}}Disaster\\ Category\end{tabular}}} & \multirow{2}{*}{\textbf{Real Image}} & \multicolumn{4}{c}{\textbf{Synthetic Image}} \\
\cmidrule{3-6}
 &  & \textbf{SD1.5} & \textbf{SD2.0} & \textbf{SDXL} & \textbf{PixArt} \\
 \midrule
Fire & 1,500 & 1,500 & 1,500 & 1,500 & 1,500 \\
Earthquake & 1,500 & 1,500 & 1,500 & 1,500 & 1,500 \\
Flood & 1,500 & 1,500 & 1,500 & 1,500 & 1,500 \\
Thunderstorm & 1,500 & 1,500 & 1,500 & 1,500 & 1,500 \\
\midrule
Total & 6,000 & 6,000 & 6,000 & 6,000 & 6,000 \\
\bottomrule
\end{tabular}
\vspace{-5mm}
\end{table}

\section{Benchmarking}

\subsection{Training and Evaluation Protocol}

To examine the model’s ability to generalize beyond its training distribution, we designed a two-stage evaluation protocol that explicitly targets out-of-distribution (OOD) detection. The task is formulated as binary classification, determining whether a given image is \textit{real} or \textit{fake}, with evaluation extending beyond in-distribution accuracy.

Our training setup isolates a single disaster category as the exclusive training domain. Real and synthetic images from this category are used to train the model, providing a controlled environment for generalization analysis. In our experiments, the \textit{earthquake} category is selected as the training domain, with all synthetic samples generated by Stable Diffusion~1.5~\cite{Rombach_2022_CVPR}. Specifically, 1,500 real and 1,500 synthetic images from the earthquake class are used for training. The models are lightly fine-tuned on this dataset to establish a consistent in-distribution baseline.

To evaluate generalization performance, two complementary OOD scenarios are defined:
\begin{itemize}
    \item \textbf{Semantic OOD (Cross-Category Evaluation):} The trained model is evaluated on the remaining three disaster categories, such as \textit{fire}, \textit{thunderstorm}, and \textit{flood}, with synthetic images generated by the same model (Stable Diffusion~1.5~\cite{Rombach_2022_CVPR}). This scenario measures the ability to generalize across unseen semantic content.
    
    \item \textbf{Generative OOD (Cross-Model Evaluation):} The model is further tested on all four disaster categories with synthetic images produced by alternative diffusion models, including Stable Diffusion~2.0~\cite{Rombach_2022_CVPR}, SDXL~\cite{podell2023sdxlimprovinglatentdiffusion}, and PixArt~\cite{chen2024pixartsigma}. This setting evaluates robustness to changes in generative mechanisms.
\end{itemize}

% \subsection{Experimental Hardware and Software}
% All neural networks used for fake image detection were implemented using standard deep learning libraries. The training and evaluation were conducted on a system equipped with two NVIDIA T4 GPUs, each with 2,560 CUDA cores and 16GB GDDR6 VRAM. The GPUs have a base clock speed of 1.59 GHz and provide sufficient compute capability for training modern convolutional and transformer-based vision models.

% All experiments were executed on a single-machine setup without distributed training. The hardware configuration was chosen to reflect accessible, cloud-based computational resources commonly used in real-world machine learning workflows.

% \section{Results and Observations}
% This section presents examples of the dataset followed by the findings of the planned computer vision experiments. The dataset is also released for the public research community for use in future studies, given the important implications of detecting AI-generated imagery.
% \footnote{The Dataset can be downloaded from: \url{https://www.kaggle.com/datasets/chritawqi/disasters-v3/}} Additionally, a supplementary dataset comprising multiple CSV files—with consistent image\_path and label fields—is provided.\footnote{The supplementary dataset is available at: \url{https://www.kaggle.com/datasets/trqduy/disaster-csv}}

\subsection{Performance of Fine-tuned Standard Deep Classifiers.}

\begin{table}[!t]
\centering
\caption{Cross-domain and cross-generator accuracy (\%) of standard classification models. Models were trained on real and SD 1.5-generated \textit{earthquake} images. The best, second-best, and third-best results are highlighted in \textbf{\textcolor{red}{bold}}, \underline{\textcolor{blue}{underline}}, and \textit{\textcolor{olive}{italic}}, respectively.}
\label{tab:trainable_results}
\resizebox{\textwidth}{!}{%
\begin{tabular}{lcccccc}
\toprule
\textbf{Test Set} & \textbf{ConvNeXt} & \textbf{Swin} & \textbf{ViT} & \textbf{SigLIPv2} & \textbf{DINOv2} & \textbf{ResNet152} \\
\midrule
\multicolumn{7}{l}{\textit{ Real Data (Unseen Categories) }} \\
Real - Fire & \underline{\textcolor{blue}{99.67}} & \underline{\textcolor{blue}{99.67}} & 88.13 & 97.87 & \textbf{\textcolor{red}{100}} & \textit{\textcolor{olive}{98.20}} \\
Real - Flood & \textbf{\textcolor{red}{99.87}} & \underline{\textcolor{blue}{99.13}} & 91.80 & \textit{\textcolor{olive}{97.67}} & 93.33 & 94.47 \\
Real - Thunderstorm & \textbf{\textcolor{red}{100}} & \underline{\textcolor{blue}{98.07}} & 70.40 & 85.93 & 95.13 & \textit{\textcolor{olive}{96.67}} \\
\midrule
\multicolumn{7}{l}{\textit{ Fake Data: Earthquake (Unseen Generators) }} \\
Earthquake - PixArt & 36.13 & \textit{\textcolor{olive}{79.17}} & \underline{\textcolor{blue}{82.79}} & 68.57 & \textbf{\textcolor{red}{89.02}} & 52.06 \\
Earthquake - SD 2.0 & 21.08 & \textit{\textcolor{olive}{82.98}} & \underline{\textcolor{blue}{95.11}} & 76.19 & \textbf{\textcolor{red}{97.46}} & 79.17 \\
Earthquake - SDXL & 35.87 & \textit{\textcolor{olive}{70.48}} & 64.57 & \underline{\textcolor{blue}{84.19}} & \textbf{\textcolor{red}{88.00}} & 45.02 \\
\midrule
\multicolumn{7}{l}{\textit{ Fake Data: Fire (Unseen Category \& Generators) }} \\
Fire - PixArt & 9.78 & \underline{\textcolor{blue}{43.87}} & \textbf{\textcolor{red}{68.00}} & \textit{\textcolor{olive}{17.40}} & 1.46 & 6.98 \\
Fire - SD 1.5 & 66.03 & \textit{\textcolor{olive}{73.27}} & \textbf{\textcolor{red}{96.83}} & \underline{\textcolor{blue}{85.71}} & 45.40 & 58.10 \\
Fire - SD 2.0 & 1.14 & 3.24 & \textbf{\textcolor{red}{94.10}} & \underline{\textcolor{blue}{55.11}} & 19.94 & \textit{\textcolor{olive}{26.35}} \\
Fire - SDXL & 11.11 & 18.29 & \textbf{\textcolor{red}{81.14}} & \underline{\textcolor{blue}{73.11}} & 8.76 & \textit{\textcolor{olive}{27.24}} \\
\midrule
\multicolumn{7}{l}{\textit{ Fake Data: Flood (Unseen Category \& Generators) }} \\
Flood - PixArt & 2.35 & \underline{\textcolor{blue}{42.86}} & \textbf{\textcolor{red}{65.97}} & 34.60 & \textit{\textcolor{olive}{35.62}} & 5.65 \\
Flood - SD 1.5 & 64.00 & 80.89 & \textbf{\textcolor{red}{99.43}} & \textit{\textcolor{olive}{85.21}} & \underline{\textcolor{blue}{92.00}} & 71.81 \\
Flood - SD 2.0 & 2.48 & 36.06 & \textbf{\textcolor{red}{97.33}} & \textit{\textcolor{olive}{63.62}} & \underline{\textcolor{blue}{78.60}} & 30.79 \\
Flood - SDXL & 3.56 & 29.21 & \textbf{\textcolor{red}{73.46}} & \underline{\textcolor{blue}{63.49}} & \textit{\textcolor{olive}{58.67}} & 19.17 \\
\midrule
\multicolumn{7}{l}{\textit{ Fake Data: Thunderstorm (Unseen Category \& Generators) }} \\
Thunderstorm - PixArt & 0.00 & \underline{\textcolor{blue}{12.23}} & \textbf{\textcolor{red}{12.79}} & 4.33 & 0.75 & \textit{\textcolor{olive}{5.02}} \\
Thunderstorm - SD 1.5 & 9.84 & \textit{\textcolor{olive}{77.78}} & \textbf{\textcolor{red}{95.30}} & \underline{\textcolor{blue}{81.90}} & 57.59 & 37.52 \\
Thunderstorm - SD 2.0 & 0.13 & 41.46 & \textbf{\textcolor{red}{93.14}} & \textit{\textcolor{olive}{55.68}} & \underline{\textcolor{blue}{58.79}} & 25.59 \\
Thunderstorm - SDXL & 0.13 & \textit{\textcolor{olive}{33.88}} & \textbf{\textcolor{red}{43.61}} & \underline{\textcolor{blue}{36.43}} & 10.87 & 19.33 \\
\bottomrule
\end{tabular}%
}
\vspace{-5mm}
\end{table}

We report the classification accuracy of various models fine-tuned on the Forged Calamity dataset. Eight vision backbones are evaluated to compare classification performance and generalization capacity: ConvNeXt~\cite{woo2023convnextv2codesigningscaling}, Swin Transformer~\cite{DBLP:journals/corr/abs-2103-14030}, ViT~\cite{wu2020visual}, SigLIPv2~\cite{zhai2023sigmoid}, DINOv2~\cite{oquab2024dinov}, and ResNet152~\cite{he2016deep}.

Results presented in Table~\ref{tab:trainable_results} confirm the central hypothesis that vision models, even after fine-tuning, exhibit a marked inability to generalize when detecting AI-generated disaster imagery. The findings reveal two critical weaknesses: poor model transferability across generators and limited domain generalization across disaster categories.

Most architectures achieve high accuracy (often above 90\%) when evaluated on images generated by Stable Diffusion~1.5~\cite{Rombach_2022_CVPR}, the model used to create the synthetic samples in the training set. However, performance declines sharply when tested on images produced by other diffusion models such as SD~2.0~\cite{Rombach_2022_CVPR}, SD~XL~\cite{podell2023sdxlimprovinglatentdiffusion}, and PixArt~\cite{chen2024pixartsigma}. This degradation indicates that detectors tend to overfit to generator-specific artifacts rather than learning generalizable cues of artificiality.

Transformer-based architectures, including Swin Transformer and ViT, generally demonstrate greater robustness and better generalization than CNN-based models such as ConvNeXt and ResNet. For instance, ConvNeXt’s accuracy drops from 98.1\% on SD~1.5 to only 49.7\% on PixArt~\cite{chen2024pixartsigma} for thunderstorm images, illustrating the brittleness of localized convolutional representations. These results suggest that the global attention mechanisms in Transformers are more effective at capturing diffuse or subtle generative artifacts that are overlooked by the limited receptive fields of CNNs.

The findings also highlight an emerging challenge: as generative models advance, their outputs become harder to detect. The notably lower accuracy on images from Stable Diffusion~XL~\cite{podell2023sdxlimprovinglatentdiffusion} suggests that artifacts in newer diffusion models are increasingly nuanced and less transferable from earlier generations. Moreover, variation in performance across disaster categories, particularly the difficulty observed with thunderstorm imagery, indicates that visual complexity and environmental detail can further impact detection reliability.

\subsection{Performance of Fine-tuned Deepfake Detectors}

\begin{table}[!t]
\centering
\caption{Cross-domain and cross-generator accuracy (\%) of fine-tuned deepfake detection models. Models were trained on real and SD 1.5-generated \textit{earthquake} images. The best, second-best, and third-best results are highlighted in \textbf{\textcolor{red}{bold}}, \underline{\textcolor{blue}{underline}}, and \textit{\textcolor{olive}{italic}}, respectively.}
\label{tab:trainable_sota_results}
\resizebox{\textwidth}{!}{%
\begin{tabular}{lccccccc}
\toprule
\textbf{Test Set} & \textbf{RINE} & \textbf{UA} & \textbf{LGrad} & \textbf{DIRE} & \textbf{FreqNet} & \textbf{ADOF} & \textbf{CGL} \\
\midrule
\multicolumn{8}{l}{\textit{ Real Data (Unseen Categories) }} \\
Real - Fire & 50.13 & \underline{\textcolor{blue}{99.53}} & 97.00 & \textit{\textcolor{olive}{99.47}} & 94.07 & 97.33 & \textbf{\textcolor{red}{100.00}} \\
Real - Flood & 50.30 & 76.00 & 89.73 & \underline{\textcolor{blue}{96.87}} & \textit{\textcolor{olive}{93.60}} & 92.73 & \textbf{\textcolor{red}{99.93}} \\
Real - Thunderstorm & 52.13 & 70.87 & 92.73 & \underline{\textcolor{blue}{98.13}} & 93.00 & \textit{\textcolor{olive}{96.60}} & \textbf{\textcolor{red}{99.00}} \\
\midrule
\multicolumn{8}{l}{\textit{ Fake Data: Earthquake (Unseen Generators) }} \\
Earthquake - PixArt & \textbf{\textcolor{red}{93.33}} & \textit{\textcolor{olive}{85.84}} & 69.84 & 49.27 & 1.59 & \underline{\textcolor{blue}{89.02}} & 82.73 \\
Earthquake - SD 2.0 & \textit{\textcolor{olive}{81.33}} & \underline{\textcolor{blue}{92.89}} & \textbf{\textcolor{red}{94.35}} & 64.89 & 25.52 & 67.05 & 11.05 \\
Earthquake - SDXL & 83.70 & \underline{\textcolor{blue}{92.57}} & \textit{\textcolor{olive}{89.40}} & 45.02 & 16.63 & \textbf{\textcolor{red}{92.76}} & 88.44 \\
\midrule
\multicolumn{8}{l}{\textit{ Fake Data: Fire (Unseen Category \& Generators) }} \\
Fire - PixArt & \textbf{\textcolor{red}{94.08}} & 43.49 & 11.37 & 10.48 & 7.62 & \textit{\textcolor{olive}{77.90}} & \underline{\textcolor{blue}{84.32}} \\
Fire - SD 1.5 & \textbf{\textcolor{red}{98.11}} & 79.11 & 60.19 & 33.14 & 75.87 & \textit{\textcolor{olive}{80.00}} & \underline{\textcolor{blue}{85.52}} \\
Fire - SD 2.0 & \textbf{\textcolor{red}{56.29}} & \underline{\textcolor{blue}{53.65}} & \textit{\textcolor{olive}{33.14}} & 2.92 & 9.02 & 5.27 & 0.70 \\
Fire - SDXL & \textbf{\textcolor{red}{64.81}} & \underline{\textcolor{blue}{58.48}} & \textit{\textcolor{olive}{46.73}} & 18.16 & 10.10 & 24.83 & 29.27 \\
\midrule
\multicolumn{8}{l}{\textit{ Fake Data: Flood (Unseen Category \& Generators) }} \\
Flood - PixArt & \textit{\textcolor{olive}{74.80}} & \textbf{\textcolor{red}{89.27}} & 34.29 & 72.38 & 11.49 & \underline{\textcolor{blue}{86.73}} & 35.75 \\
Flood - SD 1.5 & \textit{\textcolor{olive}{98.99}} & \underline{\textcolor{blue}{99.56}} & 86.67 & 90.41 & 94.86 & 93.27 & \textbf{\textcolor{red}{99.68}} \\
Flood - SD 2.0 & 67.80 & \textbf{\textcolor{red}{98.67}} & \textit{\textcolor{olive}{76.51}} & 55.81 & 41.02 & \underline{\textcolor{blue}{79.62}} & 1.46 \\
Flood - SDXL & 67.64 & \textbf{\textcolor{red}{93.46}} & \textit{\textcolor{olive}{77.27}} & 56.25 & 56.44 & \underline{\textcolor{blue}{87.81}} & 50.86 \\
\midrule
\multicolumn{8}{l}{\textit{ Fake Data: Thunderstorm (Unseen Category \& Generators) }} \\
Thunderstorm - PixArt & \textit{\textcolor{olive}{67.23}} & \underline{\textcolor{blue}{68.34}} & 5.83 & 62.07 & 5.71 & \textbf{\textcolor{red}{97.12}} & 39.06 \\
Thunderstorm - SD 1.5 & \underline{\textcolor{blue}{96.94}} & 95.68 & 63.94 & 92.51 & \textit{\textcolor{olive}{95.75}} & 94.92 & \textbf{\textcolor{red}{99.11}} \\
Thunderstorm - SD 2.0 & \textit{\textcolor{olive}{67.22}} & \textbf{\textcolor{red}{93.46}} & 61.33 & \underline{\textcolor{blue}{80.44}} & 43.75 & 65.84 & 4.44 \\
Thunderstorm - SDXL & 74.29 & 86.08 & 56.01 & \textbf{\textcolor{red}{94.15}} & 84.55 & \textit{\textcolor{olive}{88.18}} & \underline{\textcolor{blue}{93.71}} \\
\bottomrule
\end{tabular}
}
\vspace{-5mm}
\end{table}

To further validate the challenges introduced by the Forged Calamity benchmark, we extend our evaluation to include several state-of-the-art deepfake detection models specifically designed for synthetic image forensics.

We focus on data-driven deepfake detection methods built on large pre-trained vision encoders, including RINE~\cite{Koutlis:2024:RINE}, UA~\cite{Cioni:2024:UniversalAttribution}, LGrad~\cite{tan2023:Lgrad}, DIRE~\cite{wang2023dire}, FreqNet~\cite{tan:2024:freqnet}, ADOF~\cite{VoHoaiDanh:2025:ADOF}, CGL~\cite{NguyenYHop:2024:CGL}, which differ in how they extract and utilize learned representations for synthetic image detection. RINE~\cite{Koutlis:2024:RINE} and UA~\cite{Cioni:2024:UniversalAttribution} operate on intermediate-layer features of models such as CLIP~\cite{radford2021:CLIP}, under the hypothesis that these layers retain finer low-level artifacts useful for detection. In contrast, CGL~\cite{NguyenYHop:2024:CGL} also builds on CLIP but enhances performance through a caption-guided learning strategy that integrates semantic context. Moving beyond standard feature-based pipelines, LGrad~\cite{tan2023:Lgrad} introduces a gradient-based paradigm that trains a classifier directly on image gradients computed from a frozen pre-trained network. Another approach, DIRE~\cite{wang2023dire}, proposes using the Diffusion Reconstruction Error as a representation, hypothesizing that diffusion-generated images can be reconstructed more accurately by a diffusion model than real images. ADOF~\cite{VoHoaiDanh:2025:ADOF} adopts a lightweight design, applying a custom spatial high-pass filter to enhance frequency artifacts for detection with a compact CNN. Finally, representing a frequency-domain approach, FreqNet~\cite{tan:2024:freqnet} enforces spectral learning by training a compact CNN to operate on the phase and amplitude components of images.

In this experiment, each model was fine-tuned exclusively on the \texttt{earthquake} category, which included real images and synthetic counterparts generated by Stable Diffusion~1.5. The trained detectors were then evaluated on the remaining 18 unseen subsets, allowing assessment of both semantic and generative generalization across new disaster categories and diffusion models.

The results, summarized in Table~\ref{tab:trainable_sota_results}, highlight substantial generalization challenges even for these advanced detectors. CGL performs exceptionally well on unseen real images and on synthetic samples from the known generator (SD~1.5 across new categories), achieving near-perfect accuracy. However, its performance drops sharply on SD~2.0-generated images, indicating severe overfitting to generator-specific artifacts. Similarly, FreqNet fails to generalize to newer diffusion models such as SDXL and PixArt. In contrast, RINE and UA exhibit relatively stronger, though still inconsistent, generalization; for instance, RINE maintains high accuracy on SD~1.5 and PixArt “Fire” samples but struggles with SD~2.0.

These findings reinforce a consistent insight throughout our study: fine-tuning on a narrow data subset leads to strong overfitting and poor cross-domain robustness. Even models built upon powerful foundation architectures struggle to adapt to unseen generative processes, underscoring the need for broader, model-agnostic training strategies and benchmark datasets such as Forged Calamity to drive progress in reliable synthetic image detection.

\subsection{Performance of Zero-shot Generalized Deepfake Detectors}

\begin{table}[!t]
\centering
\caption{Zero-shot accuracy (\%) of generalized deepfake detection models using their publicly released pre-trained models. The best, second-best, and third-best results are highlighted in \textbf{\textcolor{red}{bold}}, \underline{\textcolor{blue}{underline}}, and \textit{\textcolor{olive}{italic}}, respectively.}
\label{tab:trainfree_sota_results}
\resizebox{\textwidth}{!}{%
\begin{tabular}{lcccccc}
\toprule
\textbf{Test Set} & \textbf{UniFD} & \textbf{DAE} & \textbf{SPAI} & \textbf{FatFormer} & \textbf{ADOF} & \textbf{CGL} \\
\midrule
\multicolumn{7}{l}{\textit{ Real Data }} \\
Real - Earthquake & 97.10 & \textbf{\textcolor{red}{100.00}} & 74.40 & \underline{\textcolor{blue}{99.53}} & 83.33 & \textit{\textcolor{olive}{99.07}} \\
Real - Fire & 88.80 & \textbf{\textcolor{red}{100.00}} & 82.80 & \textit{\textcolor{olive}{98.13}} & 85.07 & \underline{\textcolor{blue}{99.13}} \\
Real - Flood & 94.30 & \textbf{\textcolor{red}{100.00}} & 78.27 & \textit{\textcolor{olive}{98.00}} & 78.13 & \underline{\textcolor{blue}{98.73}} \\
Real - Thunderstorm & 91.60 & \textbf{\textcolor{red}{100.00}} & 70.93 & \underline{\textcolor{blue}{96.67}} & 80.47 & \textit{\textcolor{olive}{92.80}} \\
\midrule
\multicolumn{7}{l}{\textit{ AI Data: PixArt }} \\
PixArt - Earthquake & 2.10 & 67.75 & \textit{\textcolor{olive}{70.98}} & 5.97 & \textbf{\textcolor{red}{99.17}} & \underline{\textcolor{blue}{82.73}} \\
PixArt - Fire & 6.00 & 73.52 & \underline{\textcolor{blue}{94.16}} & 5.02 & \textbf{\textcolor{red}{99.81}} & \textit{\textcolor{olive}{84.32}} \\
PixArt - Flood & 2.00 & \textit{\textcolor{olive}{70.98}} & \underline{\textcolor{blue}{96.63}} & 8.83 & \textbf{\textcolor{red}{99.68}} & 35.75 \\
PixArt - Thunderstorm & 4.80 & \textit{\textcolor{olive}{59.56}} & \underline{\textcolor{blue}{96.49}} & 13.86 & \textbf{\textcolor{red}{100.00}} & 39.06 \\
\midrule
\multicolumn{7}{l}{\textit{ AI Data: Stable Diffusion 1.5 }} \\
SD 1.5 - Earthquake & 9.10 & 64.70 & \textbf{\textcolor{red}{100.00}} & 27.62 & \underline{\textcolor{blue}{99.81}} & \textit{\textcolor{olive}{87.62}} \\
SD 1.5 - Fire & 17.90 & \underline{\textcolor{blue}{96.51}} & \textbf{\textcolor{red}{99.94}} & 24.95 & \textit{\textcolor{olive}{95.17}} & 69.78 \\
SD 1.5 - Flood & 8.30 & \textit{\textcolor{olive}{90.16}} & \textbf{\textcolor{red}{100.00}} & 36.70 & \underline{\textcolor{blue}{99.68}} & 74.35 \\
SD 1.5 - Thunderstorm & 17.30 & \textit{\textcolor{olive}{95.24}} & \underline{\textcolor{blue}{99.75}} & 50.92 & \textbf{\textcolor{red}{100.00}} & 80.25 \\
\midrule
\multicolumn{7}{l}{\textit{ AI Data: Stable Diffusion 2.0 }} \\
SD 2.0 - Earthquake & 0.30 & \textbf{\textcolor{red}{100.00}} & \textit{\textcolor{olive}{92.06}} & 4.83 & \underline{\textcolor{blue}{97.78}} & 11.05 \\
SD 2.0 - Fire & 11.00 & \textbf{\textcolor{red}{99.11}} & \underline{\textcolor{blue}{87.68}} & 0.19 & \textit{\textcolor{olive}{29.02}} & 0.70 \\
SD 2.0 - Flood & 0.30 & \textbf{\textcolor{red}{100.00}} & \textit{\textcolor{olive}{97.08}} & 6.35 & \underline{\textcolor{blue}{99.17}} & 1.46 \\
SD 2.0 - Thunderstorm & 3.80 & \underline{\textcolor{blue}{98.16}} & \textit{\textcolor{olive}{89.59}} & 11.17 & \textbf{\textcolor{red}{99.43}} & 4.44 \\
\midrule
\multicolumn{7}{l}{\textit{ AI Data: Stable Diffusion XL }} \\
SDXL - Earthquake & 4.00 & 68.38 & \underline{\textcolor{blue}{98.86}} & 23.81 & \textbf{\textcolor{red}{99.94}} & \textit{\textcolor{olive}{88.44}} \\
SDXL - Fire & 6.00 & \textit{\textcolor{olive}{34.67}} & \underline{\textcolor{blue}{95.17}} & 11.05 & \textbf{\textcolor{red}{99.24}} & 29.27 \\
SDXL - Flood & 1.70 & \textit{\textcolor{olive}{55.75}} & \underline{\textcolor{blue}{99.68}} & 33.59 & \textbf{\textcolor{red}{100.00}} & 50.86 \\
SDXL - Thunderstorm & 6.70 & 56.52 & \underline{\textcolor{blue}{97.58}} & 28.61 & \textbf{\textcolor{red}{100.00}} & \textit{\textcolor{olive}{93.71}} \\
\bottomrule
\end{tabular}%
}
\vspace{-5mm}
\end{table}

The zero-shot evaluation assesses the ability of detection models to distinguish real from synthetic images without any task-specific training on the Forged Calamity dataset. This scenario evaluates the baseline robustness of each model’s underlying mechanism. We evaluated six state-of-the-art generalized deepfake detection methods: UniFD~\cite{ojha2023:universalFakeDetect:UniFD}, DAE~\cite{vesnin:2024:detectingAE}, SPAI~\cite{karageorgiou2025:SPAI}, FatFormer~\cite{liu2024:FatFormer}, the zero-shot versions of ADOF~\cite{VoHoaiDanh:2025:ADOF} and CGL~\cite{NguyenYHop:2024:CGL}. We used their publicly available pre-trained models on other deepfake detection datasets without any fine-tuning.

UniFD~\cite{ojha2023:universalFakeDetect:UniFD} adopts a simple approach by training a linear classifier on frozen features extracted from CLIP. DAE~\cite{vesnin:2024:detectingAE} represents an architecture-specific approach designed for LDMs. It detects artifacts introduced by the Autoencoder (AE) component, which often leave subtle structural traces in generated images. SPAI~\cite{karageorgiou2025:SPAI} models the spectral distribution of real images through a self-supervised frequency reconstruction task, identifying synthetic content as out-of-distribution samples. FatFormer~\cite{liu2024:FatFormer} introduces an adaptive transformer framework that fine-tunes a foundation model using a forgery-aware adapter and a language-guided contrastive objective to learn more resilient forgery representations.

The results, summarized in Table~\ref{tab:trainfree_sota_results}, show that ADOF and SPAI achieve the highest overall performance in this challenging setting. ADOF attains near-perfect accuracy on images generated by advanced diffusion models such as SDXL and PixArt, demonstrating the effectiveness of its frequency-based filtering approach. DAE also performs strongly, perfectly identifying real images and excelling on SD~2.0 samples, consistent with its design to detect AE-related artifacts. However, no method performs uniformly across all conditions. ADOF’s accuracy declines noticeably on SD~2.0 “Fire” images and trails DAE in identifying real samples. In contrast, FatFormer consistently underperforms, systematically failing to identify fakes, indicating that its adaptive fine-tuning mechanism is poorly suited for diverse, high-fidelity imagery without dataset-specific adaptation.

Overall, these findings confirm that generalization remains a major challenge in synthetic image detection. The Forged Calamity benchmark clearly highlights the strengths and weaknesses of current detection methods and provides a valuable foundation for advancing more universal and reliable fake image detectors.

\subsection{Failure Cases}
\begin{figure}[!t]
    \centering
    \includegraphics[width=0.85\textwidth]{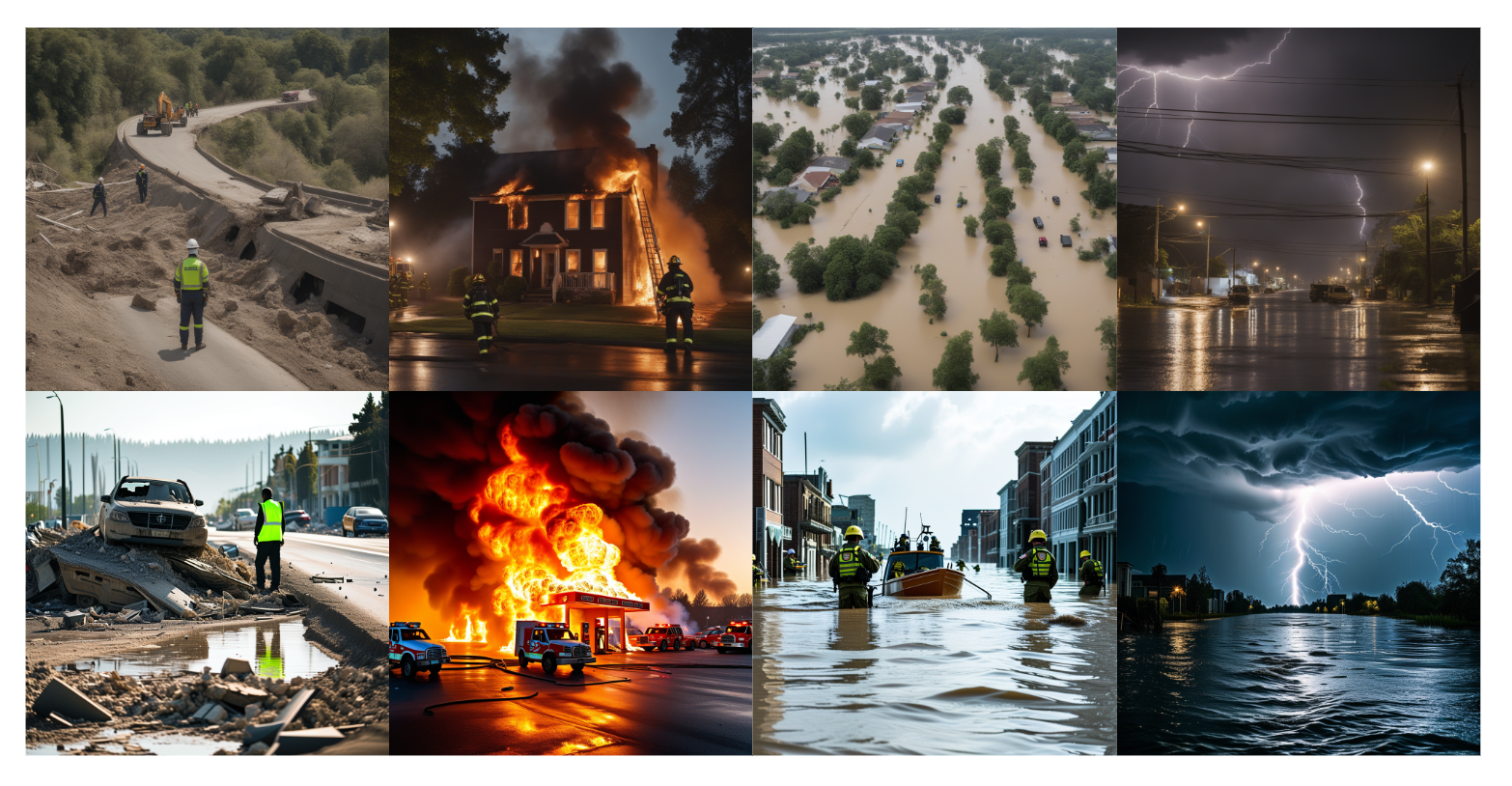}
    \caption{Failure cases, tested on SigLipv2.}
    \label{fig:failure_cases}
    \vspace{-5mm}
\end{figure}
As shown in Figure~\ref{fig:failure_cases}, models like SigLipv2 still exhibit certain limitations, mainly because they may not fully capture the distribution of images generated by diffusion models, nor accurately model the lighting composition and fine details that differ from real-world scenes. This likely results in misclassification when synthetic outputs closely resemble real photographs in terms of texture and overall realism. A similar tendency can be observed in other detectors, suggesting that these models may not possess a sufficiently holistic understanding of both global and local image information. They often fail to preserve coherent color composition and tend to overlook subtle local anomalies, such as irregular text or unnatural object details, which ultimately prevents them from reliably distinguishing fake images from real ones.

\section{Conclusion}

In this paper, we addressed the critical challenge of detecting synthetic disaster imagery by systematically evaluating the generalization capabilities of modern vision models. We introduced Forged Calamity, a large-scale benchmark dataset consisting of 30,000 images, including 6,000 real and 24,000 synthetic samples generated from four diffusion models across four disaster categories. The dataset provides a comprehensive and balanced testbed for analyzing the generalization and resilience of fake image detectors in high-stakes contexts such as disaster response and media forensics. 

Through extensive experiments under both fine-tuned and zero-shot conditions, we observed that most existing approaches achieve strong accuracy within their training distributions but struggle significantly when evaluated on unseen diffusion models or new disaster categories. Fine-tuned detectors tend to overfit to generator-specific artifacts, with accuracy reductions of up to 50\% in cross-generator evaluations. Zero-shot models show partial robustness yet fail to maintain consistent accuracy across diverse generative sources, confirming the limited generalization capacity of current detection strategies.

These findings reveal that the majority of existing forensic models remain heavily dependent on model-specific visual cues and are not yet capable of reliably distinguishing real from synthetic content in open-world conditions. As diffusion models continue to evolve and produce more realistic outputs with minimal detectable artifacts, this limitation becomes increasingly critical for digital forensics, cybersecurity, and misinformation prevention.

In future work, we plan to expand Forged Calamity to include additional generative models, real-world social media imagery, and multimodal metadata to better represent realistic misinformation scenarios. We will also explore domain-agnostic and cross-modal detection frameworks that integrate reasoning, visual-language alignment, and explainable AI to develop more generalizable and trustworthy detection systems capable of safeguarding visual authenticity in the diffusion era.

\begin{credits}
\subsubsection{\ackname} 
This research is funded by Vietnam National University - Ho Chi Minh City (VNU-HCM) under Grant Number C2024-18-25. This research used the GPUs provided by the Intelligent Systems Lab at the Faculty of Information Technology, University of Science, VNU-HCM.

Prof. Isao Echizen is supported by JSPS KAKENHI Grant JP24H00732, by JST CREST Grants JPMJCR20D3 and JPMJCR2562 including AIP challenge program, and by JST K Program Grant JPMJKP24C2 Japan.
\end{credits}

\bibliographystyle{splncs04}
\bibliography{ref}

\end{document}